\documentclass[conference]{IEEEtran}
\usepackage{cite}
\usepackage{amsmath,amssymb,amsfonts}
\usepackage{algorithmic}
\usepackage{graphicx}
\usepackage{textcomp}
\usepackage{xcolor}
\usepackage{caption}
\usepackage{subcaption}

\def\BibTeX{{\rm B\kern-.05em{\sc i\kern-.025em b}\kern-.08em
    T\kern-.1667em\lower.7ex\hbox{E}\kern-.125emX}}
\begin{document}

\title{Private data sharing between decentralized users through the privGAN architecture}

\author{\IEEEauthorblockN{1\textsuperscript{st} Jean-Francois Rajotte}
\IEEEauthorblockA{\textit{Data Science Institute} \\
\textit{University of British Columbia}\\
Vancouver, Canada \\
jfraj@mail.ubc.ca}
\and
\IEEEauthorblockN{2\textsuperscript{nd} Raymond T. Ng}
\IEEEauthorblockA{\textit{Department of Computer Science} \\
\textit{University of British Columbia}\\
Vancouver, Canada \\
rng@cs.ubc.ca}
}

\maketitle

\begin{abstract}
More data is almost always beneficial for analysis and machine learning tasks.
In many realistic situations however, an enterprise cannot share its data, either to keep a competitive advantage or to protect the privacy of the data sources, the enterprise's clients for example.
We propose a method for data owners to share synthetic or fake versions of their data without sharing the actual data, nor the parameters of models that have direct access to the data.
The method proposed is based on the privGAN architecture where local GANs are trained on their respective data subsets with an extra penalty from a central discriminator aiming to discriminate the origin of a given fake sample.
We demonstrate that this approach, when applied to subsets of various sizes, leads to better utility for the owners than the utility from their real small datasets.
The only shared pieces of information are the parameter updates of the central discriminator.
The privacy is demonstrated with white-box attacks on the most vulnerable elments of the architecture and the results are close to random guessing.
This method would apply naturally in a federated learning setting.
\end{abstract}


\section{Introduction}

\begin{figure}
\centering\includegraphics[width=7cm]{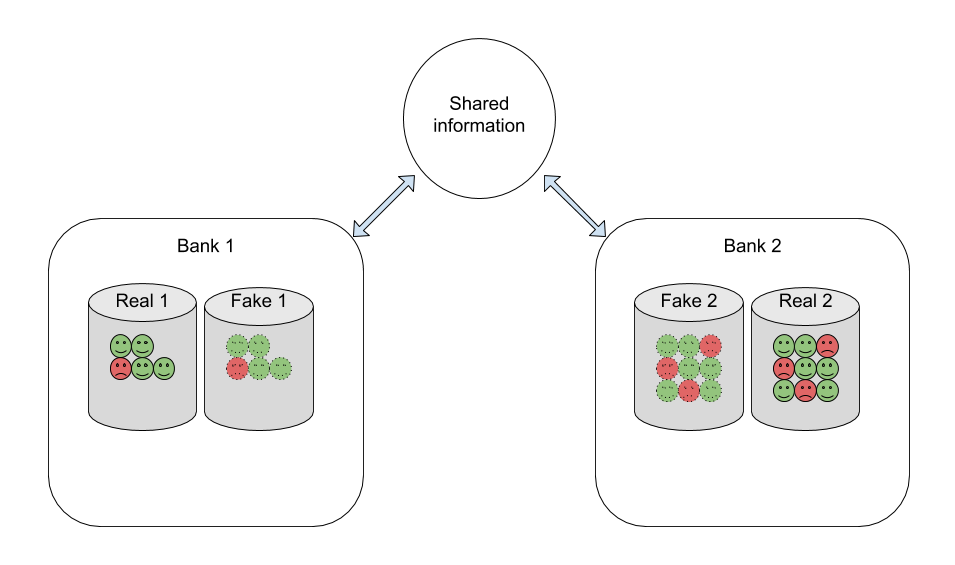}
\caption{\em A motivating example: conceptual representation of information sharing with synthetic data from two banks.
The data samples here are represented by green smileys and red frownies, which could represent the good and fraudulent customers signatures respectively.
Banks would benefit to learn from more fraudulent signatures but cannot share their data.
The open question is what the banks can share to improve their knowledge of fraudulent users while minimizing the risk of privacy breaches of their customers.}
\label{fig:bank_sharing}
\end{figure}

Sharing data without breaching privacy is a common challenge in many fields because of lengthy approval processes required to address ethical and privacy requirements. 
As a motivating example, one can imagine two banks (or two bank branches) who would like to build a model to detect fraudulent handwritten signatures but have too few samples on their own to train a sufficiently accurate fraudulent behaviour classification model.
There are many ways to solve this problem; but the situation can be generally represented by Fig. \ref{fig:bank_sharing}. The crux of the issue is  the exact definition of \textit{Shared Information}. Specifically, the two banks have at least the following options with respect to what is to be shared:

\begin{itemize}
    \item their data;
    \item synthetic data generated by a basic Generative Adversarial Network (GAN); or
    \item access to synthetic samples while the generator is being trained.
\end{itemize}
We assume in this paper that the first option is too risky; but we will keep it as an ideal baseline that we want to be as close as possible.

A natural method for using data from multiple sources while keeping data locally is federated learning \cite{mcmahan2016communicationefficient}; but that would allow data access to model(s) shared among data owners.
One could add a layer of security by creating a synthetic data either locally or centrally with GANs \cite{NIPS2014_5423}.
For the banks example, the fraudulent signatures classifier could be trained on the synthetic data.
In their typical implementations \cite{augenstein2019generative, hardy2018mdgan, rasouli2020fedgan}, however, GANs trained with a federated learning framework necessitate sharing a model that has direct access to the data owners' (e.g. the banks) data, which is again a privacy risk.

Our approach studied here is a method that generates local synthetic samples that comprise characteristics from all data owners' data while {\em only sharing a central discriminator that access synthetic samples}.
Motivated by real world use cases such as the banks in Fig. \ref{fig:bank_sharing}, we modified the original privGAN \cite{mukherjee2019privGAN} prescription to create disjoint subsets of varying sizes.
Another notable difference with the original privGAN is the exclusion of a pre-training of the central discriminator on the real data since the whole point of this work is to circumvent the real data access limitations.

As a proxy for the client signatures, we use the hand written digits from which we generate synthetic fake digits.
Like the original privGAN, we show that the resulting local synthetic data are  more private than local synthetic data resulting from local GANs. In order to demonstrate the validity of this approach, 
we will consider utility as the classification accuracy of a model trained on the synthetic digits and evaluated on a holdout set.

\section{Related works}
Sharing private data or their characteristics has been extensively explored recently. A very natural approach is simply to generate private synthetic data with Differential Privacy \cite{10.1561/0400000042} through GANs based on either differentially private stochastic gradient descent \cite{Abadi_2016} (e.g \cite{xie2018differentially}) or the PATE \cite{papernot2016semisupervised} (e.g. \cite{yoon2018pategan}). Both approaches suffer from low utility data for a reasonable degree of privacy. A recent paper trying to address this issue shows some promise \cite{neunhoeffer2020private}.

Another approach is to train a model in a federated learning setting such that the data never has to be shared; some examples are FedGAN \cite{rasouli2020fedgan}, MD-GAN \cite{hardy2018mdgan} and \cite{fan2020federated}.
Since it has been demonstrated that GANs are vulnerable to privacy attacks \cite{DBLP:journals/corr/HayesMDC17}, various ways have been proposed to provide better privacy protection.
GANs trained on distributed datasets with differential privacy (e.g. \cite{geyer2017differentially} and \cite{chen2020gswgan}) suffer from the same low quality generated samples as centrally trained GANs, unless they have access to a very large amount of training data as in this language model application \cite{mcmahan2017learning}.
FedGP \cite{triastcyn2019federated} addresses these challenges by training a central generator while keeping the discriminators and their parameters local.
The privacy protection, however, is obtained through Differential Average-Case Privacy using "post-hoc privacy analysis framework". Our method allows users to create local synthetic data sets, and privacy protection naturally arises from the architecture without post processing steps.

\section{privGAN}
\begin{figure}
\centering\includegraphics[width=5cm]{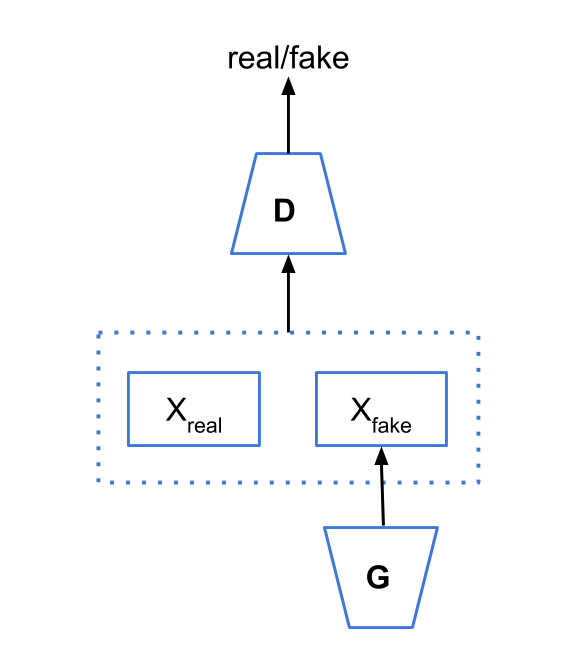}
\caption{\em Simple GAN architecture:
the generator \textbf{G} aims to create fake samples \emph{X\textsubscript{fake}} that are indistinguishable from the real samples \emph{X\textsubscript{real}} by the discriminator \textbf{D}.
}
\label{fig:simple_gan}
\end{figure}

PrivGAN\cite{mukherjee2019privGAN} is an extension of GAN whose development has lead to notoriously realistic synthetic images.
In their original version, GANs comprise a generator \textbf{G} and a discriminator \textbf{D} playing a two-player game as shown in Fig. \ref{fig:simple_gan}.
The generator aims to create fake samples such that the discriminator estimates their probability to be real as high as possible.
The discriminator on the other hand, tries to estimate the probability that a sample is real rather than fake.

\begin{figure}
\centering\includegraphics[width=7cm]{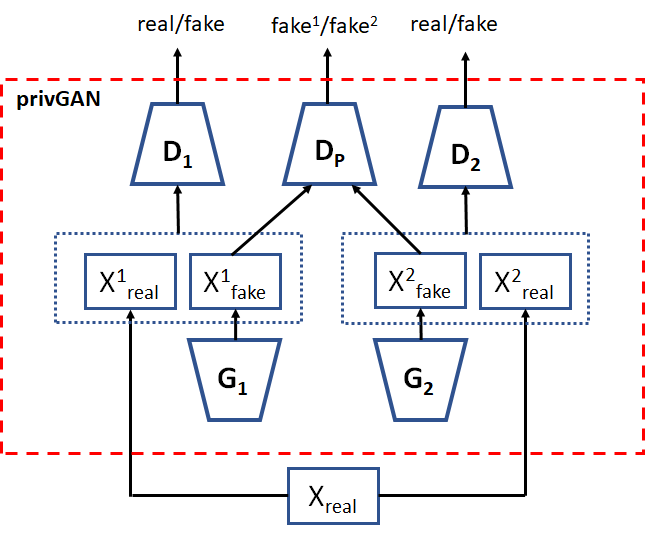}
\caption{\em Original privGAN architecture with N=2 subsets.
In relation to Fig. \ref{fig:bank_sharing}, the \textit{shared information}
corresponds to the central discriminator \emph{\textbf{D\textsubscript{P}}} accessing only the fake samples during the training.}
\label{fig:privGAN_original}
\end{figure}

PrivGAN was originally designed to protect against membership inference attacks such as \cite{DBLP:journals/corr/HayesMDC17}.
The architecture comprised $N$ GANs trained on their disjoints, independent and identically distributed (i.i.d.) subsets with an extra loss from a central (private) discriminator \textbf{D\textsubscript{P}} as described in Fig. \ref{fig:privGAN_original}.
The numeric subscripts correspond to the models and data associated to a given data owner.
In the banks example from Fig. \ref{fig:bank_sharing}, $X_1$ corresponds to Bank 1's clients' signatures and \textbf{G\textsubscript{1}}-\textbf{D\textsubscript{1}} correspond to the GAN's generator-discriminator pair trained on those signatures with the extra discriminator \textbf{D\textsubscript{P}} shared by both banks.
The authors show that their method ``\emph{minimally affects the quality of downstream samples as evidenced by the performance on downstream learning tasks such as classification}''.
For the banks, the two important things in this setting are 1) the clients' signatures never leave the bank and 2) architecture elements that have direct access to the real signatures never leave the bank either.

In this paper, we explore the application of the privGAN architecture on subsets that are different in sizes.
Our exploration will focus on the utility of the synthetic dataset as a training set for a classification model which we evaluate on a holdout set from the original data.
Our only other modification of the original method is to skip the pre-training of the central discriminator \textbf{D\textsubscript{P}} on real data.
This is to limit real data access to local components.
In our bank example from Fig. \ref{fig:bank_sharing}, that means that the client signatures from Bank 1 are only accessed by the discriminator \textbf{D\textsubscript{1}} which itself is never shared with the other bank.
We will see that reducing Bank 1's amount of signatures eventually leads to significant reduction of the utility of the synthetic data even to a point of uselessness.
We will show how Bank 2 could help Bank 1 without compromising it's client's privacy.

\section{Experimental details}

In our experiments, we compare the utility and privacy of the fake data generated with privGAN and the simple (non-private) GAN.
We limit our exploration to two data owners (e.g. two banks) so the architecture is exactly as depicted in Fig. \ref{fig:privGAN_original}.
As a proxy for the banks' client hand written signatures, we use MNIST \cite{lecun2010mnist} to demonstrate our approach.
MNIST contains 70,000 (60,000 training samples and 10,000 test samples) grayscale handwritten digits.
It contains a balanced number of each 10 classes (digits from 0 to 9) of 28x28 images.
We vary both training subsets in size to cover various scenarios and are referred to as a fraction of the total subset.
For examples, we show results for a data owners has as low as 1\% of the total training dataset (60,000 images) corresponding to 600 images, i.e. 60 images per digit.
In the privGAN architecture, we combine this data owner with another one also with varying dataset size.
We explore combinations of data owners with various fractions from 1\%-1\% to 50\%-50\%\footnote{We make the maximum data in a privGAN subset to be 50\% of the dataset which is more than enough to reach very high utility.} and many in between such as 20\%-10\%, 8\%-50\%, etc.
In our bank analogy, this corresponds to various scenarios where the two banks have different amounts of signatures to train their synthetic generator.

\subsection{Models and training}
As for the original privGAN paper, we use standard fully connected networks for both generators and discriminators because MNIST is a relatively simple dataset.
Identical generator and discriminator architectures are used for both GANs and privGANs.
We train all the privGAN models with Adam optimizer with learning rate of 0.0002 ($\beta$=0.5) for 200 epochs.
We use the same parameters for the GAN models except for 250 epochs.
Following the procedure of the original paper, we run a model for each digit in order to have a label for each sample.
As mentioned before, contrary to the original privGAN method, we do not pre-train the central discriminator \textbf{D\textsubscript{P}} on real images.
It is not expected to have a significant impact as this pre-training was only implemented to speed up the convergence in the original privGAN paper.\\

While evaluating the performance on downstream classification tasks, we train all GAN models with an Adam optimizer with a learning rate of 0.0002 ($\beta$=0.5) for 25 epochs over 3000 generated samples\footnote{When evaluating privGAN generated samples from both generators, the evaluation is still performed on 3000 samples, 1500 from each generators.}.
In all cases, we use batch size of 256 samples.

\subsection{Membership inference attacks}
To evaluate the privacy vulnerability of a model and its generated data, we apply membership attacks on the discriminators, the most vulnerable elements of the architecture.
The white-box attacks on the simple GANs are outlined in \cite{hayes2017logan} where it is assumed that the adversary is in possession of the trained model along with a large data pool including the training samples.
The attacker is also assumed to have the knowledge of what fraction of the dataset was used for training, $f$.
The attack then proceeds by using the discriminator of the trained GAN to obtain a probability score for each sample in the dataset.
The samples are then sorted in descending order of probability score and the top $f$ fraction of the scores are predicted as the likely training set.
The evaluation of the white-box attack is done by calculating what fraction of the predicted training set was actually in the training set.

In our bank example, a successful membership inference attack would trace back clients to their bank, an information to be kept private.
It makes the reasonable assumption that the adversary has access to clients' signatures which can come from many sources (e.g. contracts or petitions signed by the client).\\

Since our privGAN models have two generator–discriminator pairs, the previously described attack cannot be directly applied to it.
However, for a successful white–box attack, each of the discriminators should score samples from the training corpus higher than those not used in training.
The privGAN authors modified the previous approach by identifying a ’max’ probability score by taking the max over the scores from all discriminators.
The rationale being that the discriminator which has trained on a particular data sample should have the largest discriminator score.
We now proceed to sort the samples by each of these aggregate scores and select the top $f$ fraction samples as the predicted training set.
As a sanity check, we also include the mean probability score in our results.

\section{Results}

\begin{figure}
\centering\includegraphics[width=0.5\textwidth]{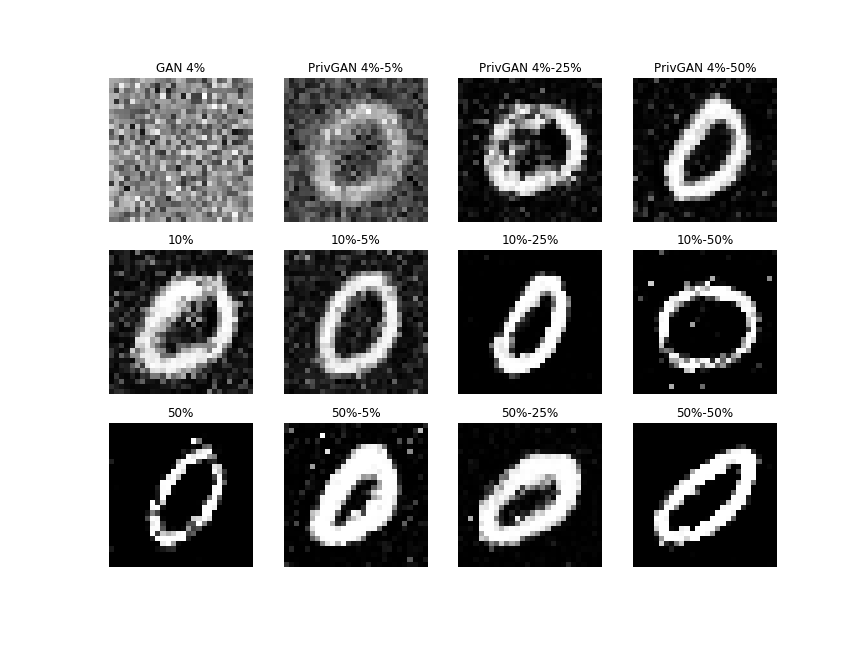}
\caption{\em Generated images with GANs (left column) and privGAN's generator 1 (other columns).
The rows corresponds to the fraction of the training data accessible locally.
For the GAN, this correspond to the total training data size and for privGAN, this corresponds to the subsets $X_{1}^{real}$.
Column 2,3,4 correspond to the data size of the other data owner, $X_{2}^{real}$.}
\label{fig:synthetic_mnist}
\end{figure}

We first look at the generated images from a GAN and privGAN trained on various training set sizes in Fig. \ref{fig:synthetic_mnist}.
One can see that for small training data (4\%-5\% of the training set), GANs are just creating noise.
PrivGAN on the other hand, does create recognizable digits even when the other subset is also small (shown at 4\% in the figure).
We also note that when a data owner has a large amount of data, privGAN makes the image noisier than what a simple GAN would generate.
This is not necessary a bad thing, it can be interpreted as a sign of privacy.
Indeed, going back to our bank analogy, a noisier synthetic signature is harder to link to any real signature from the bank's clients.
We will see below that this privacy comes at a low cost in utility for synthetic samples created with privGAN.\\

\begin{figure}
     \centering
     \begin{subfigure}[b]{0.5\textwidth}
         \centering
         \includegraphics[width=\textwidth]{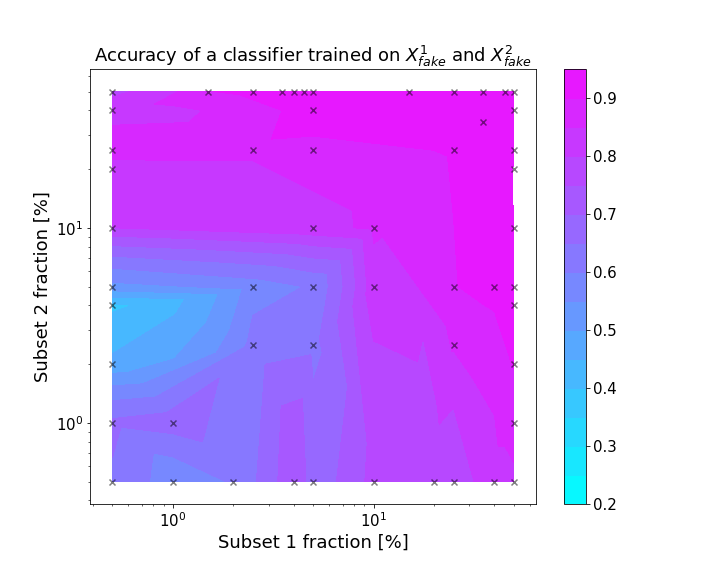}
         \caption{Trained on samples from both generators}
         \label{fig:accuracy_2d_bothgen}
     \end{subfigure}
     \hfill
     \begin{subfigure}[b]{0.5\textwidth}
         \centering
         \includegraphics[width=\textwidth]{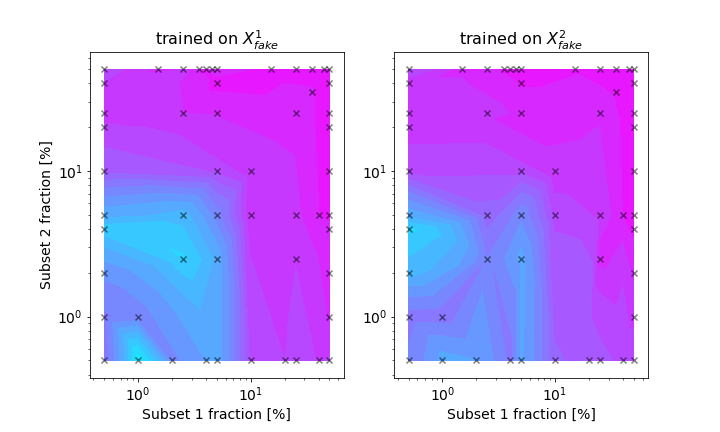}
         \caption{Trained on samples from generator 1 (left) and 2 (right)}
         \label{fig:accuracy_2d_bygen}
     \end{subfigure}
        \caption{\em Accuracy (as color) of a classification model trained on privGAN synthetic data.
        All figures share the same color code.
        The vertical and horizontal axes correspond to the fraction of data available for training the generator-discriminator pair.
        The fraction is w.r.t. the full training set for each digit, i.e. 100\% = 10,000 training samples per digit.
        The black points correspond to actual accuracies and the color filling between the points are interpolations. Note the log scale on the horizontal and vertical axes.}
        \label{fig:accuracy_2d}
\end{figure}

\begin{figure}
\centering\includegraphics[width=7cm]{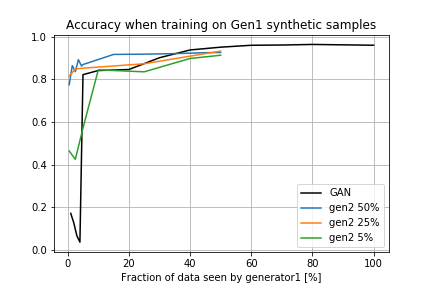}
\caption{\em Accuracy of a classification model trained on privGAN and GAN synthetic data.
The privGAN synthetic data is comprised of samples from generator 1 only.
The horizontal axes correspond to the fraction of the training sample subset used for training the privGAN generator-discriminator pair while keeping the other subset size fixed.}
\label{fig:accuracy_1d_with_gan}
\end{figure}

We then explore the effect on the utility of an unequal amount of data on each data owners.
The subsets are i.i.d. samples of different sizes from the parent training population so the data distributions are the same, only the subset sizes vary.
Fig. \ref{fig:accuracy_2d} shows the classification accuracy of a model trained separately on each local synthetic dataset and their union.
For reference, the point 50\%-50\% on the most top right of Fig. \ref{fig:accuracy_2d_bothgen} correspond to the original privGAN paper where the whole training set was used.
We note that, for the privGAN architecture, the data owners are interchangeable which translate in the symmetry along a bottom-left to top-right line which is reasonably respected within some fluctuations.
For example, the accuracy at 10\%-50\% should be the same as 50\%-10\%.
We further note that the utility remains high for both synthetic datasets as long as one has sufficiently enough training data.
This can be seen at the lower-right and upper-left sections of Fig. \ref{fig:accuracy_2d_bygen}.

Fig. \ref{fig:accuracy_1d_with_gan} compares the same accuracies with those for the same classifier trained on synthetic samples from a local GAN.
We first notice that at some threshold (5\%), GAN produce synthetic data with essentially no utility: the classifier accuracy is close to random guessing.
This corresponds to GAN generating noise as noted in Fig. \ref{fig:synthetic_mnist}.
This does not happen for privGAN, there is always a minimal utility observed in our exploration.
Clearly, privGAN is beneficial when data owners have small datasets size even if the other data owner has small data size too.
For the bank analogy, this means that banks can help each others in building better signature classification models even when they have few signatures.
As the available training data size increase, all utilities converge slightly below the GAN utility.
Indeed, when the subset reaches the size of 40\%, there is no utility gain from the privGAN architecture (although we will see below there is still a privacy benefit).
There is always a subset size threshold, depending on the other data owner subset size, below which privGAN becomes beneficial in utility.
For example, when the other data owner has as a 25\% subset size, privGAN becomes beneficial for subsets size below roughly 20\%.
\\

\begin{figure}
\centering
\begin{subfigure}{0.2\textwidth}
  \centering
         \includegraphics[width=\textwidth]{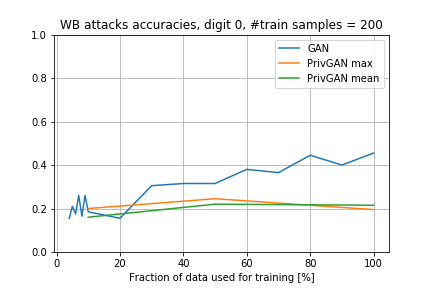}
\end{subfigure}%
\begin{subfigure}{0.2\textwidth}
  \centering
  \includegraphics[width=\textwidth]{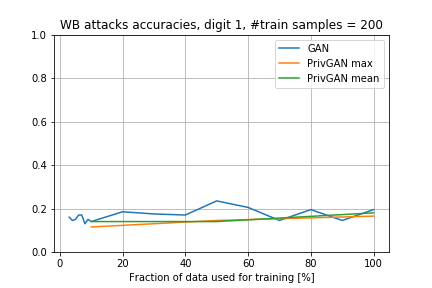}
\end{subfigure}
\begin{subfigure}{0.2\textwidth}
  \centering
  \includegraphics[width=\linewidth]{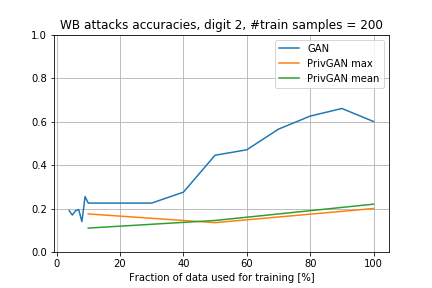}
\end{subfigure}%
\begin{subfigure}{0.2\textwidth}
  \centering
  \includegraphics[width=\linewidth]{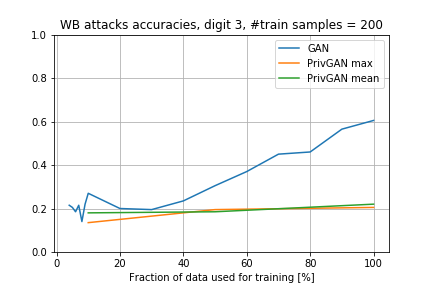}
\end{subfigure}
\caption{\em White-box attack accuracies as function of fraction of training data used for training privGAN. Here, both subsets have the same fraction of the data, so 100\% corresponds to each subset having 50\% of the full training data set.
The attacks are performed with 200 training samples combined with the X (approx 1000) hold out from the given digit which makes the random guessing at 0.2.
All privGAN attacks are close to random guessing while most GAN results are above and increasing with data fraction.
}
\label{fig:wbattacks}
\end{figure}

\begin{figure}
\centering
\begin{subfigure}{0.5\textwidth}
  \centering
  \includegraphics[width=0.55\textwidth]{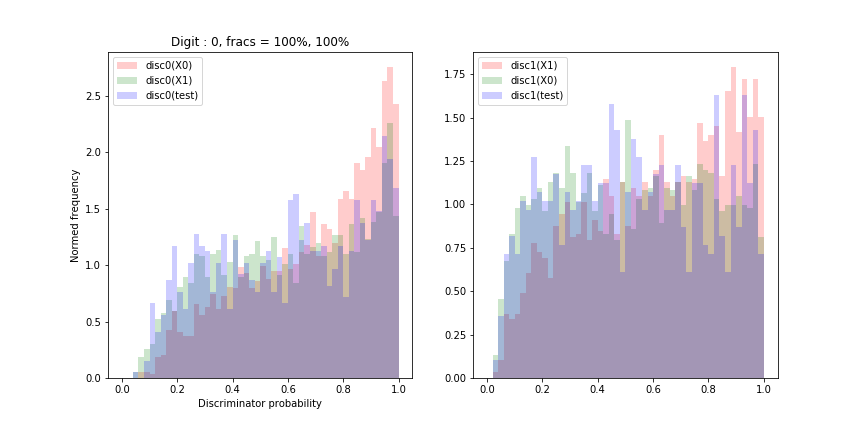}
  \includegraphics[width=0.4\textwidth]{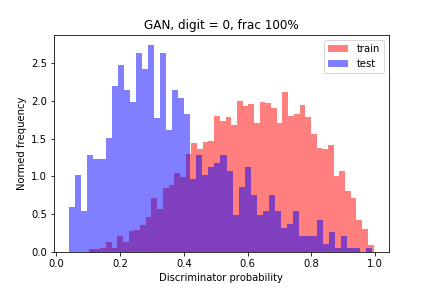}
\end{subfigure}
\begin{subfigure}{0.5\textwidth}
  \centering
  \includegraphics[width=0.55\textwidth]{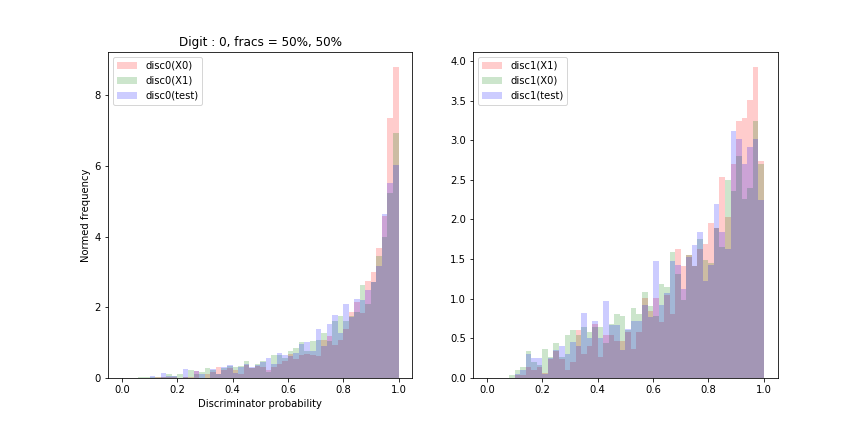}
  \includegraphics[width=0.4\textwidth]{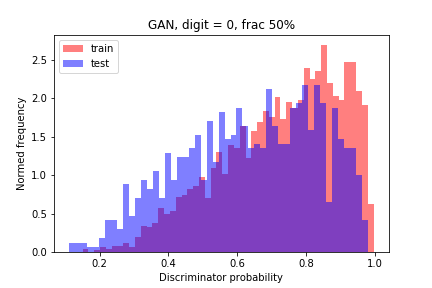}
\end{subfigure}
\begin{subfigure}{0.5\textwidth}
  \centering
  \includegraphics[width=0.55\textwidth]{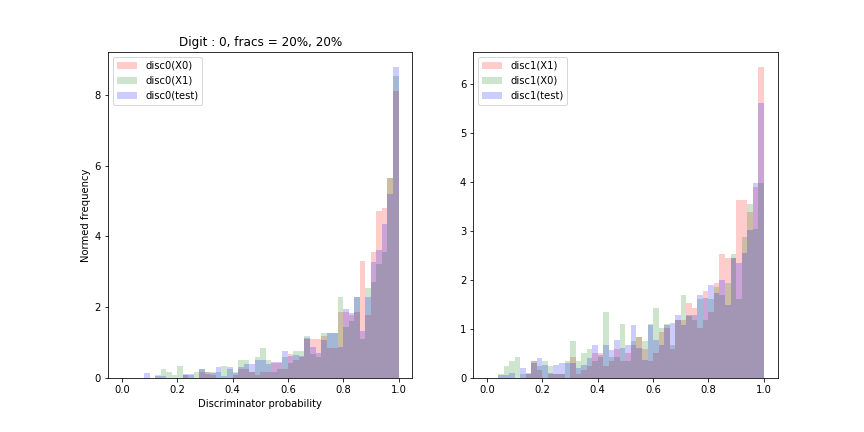}
  \includegraphics[width=0.4\textwidth]{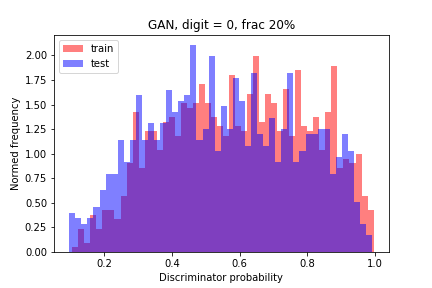}
\end{subfigure}
\caption{\em Digit 0 discriminator distributions.
The rows correspond to subsets fraction = 100\%, 50\% \& 20\% of the training data, respectively.
The columns correspond to the discriminator 1, discriminator 2 \& GAN discriminator, respectively.
The red color corresponds to the discriminator probability on its training data,
the blue color corresponds to the hold out data and the green color correspond to the training data of the other discriminator (privGAN only).
This shows how the privGAN discriminators give similar probabilities for train and test set, a characteristic of private models.
For GANs, there is a clear distinction of the distributions.}
\label{fig:disc_dist_digit0}
\end{figure}

\begin{figure}
\centering
\begin{subfigure}{0.5\textwidth}
  \centering
  \includegraphics[width=0.55\textwidth]{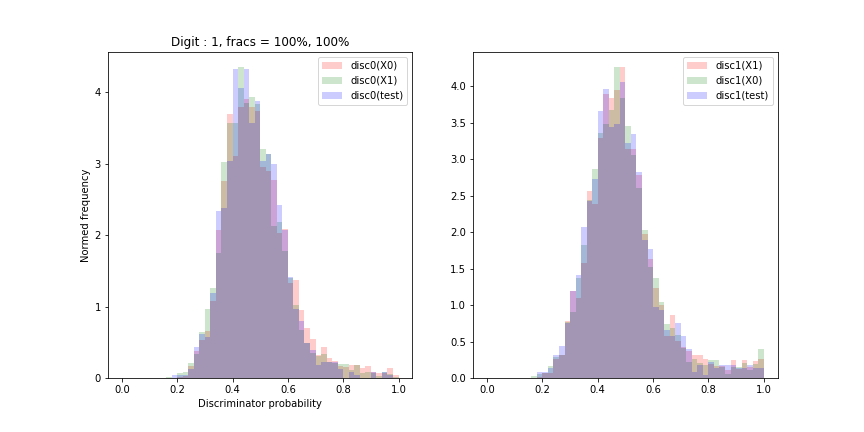}
  \includegraphics[width=0.4\textwidth]{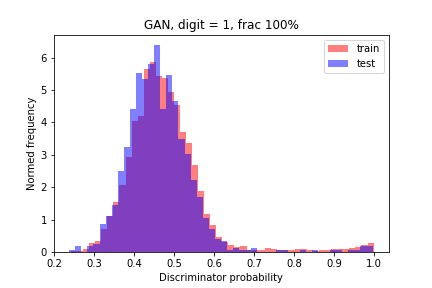}
\end{subfigure}
\begin{subfigure}{0.5\textwidth}
  \centering
  \includegraphics[width=0.55\textwidth]{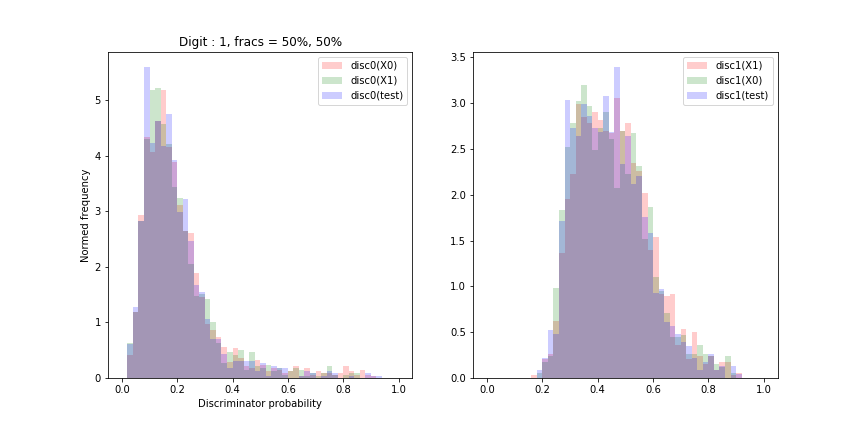}
  \includegraphics[width=0.4\textwidth]{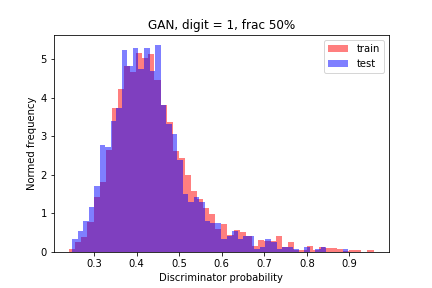}
\end{subfigure}
\begin{subfigure}{0.5\textwidth}
  \centering
  \includegraphics[width=0.55\textwidth]{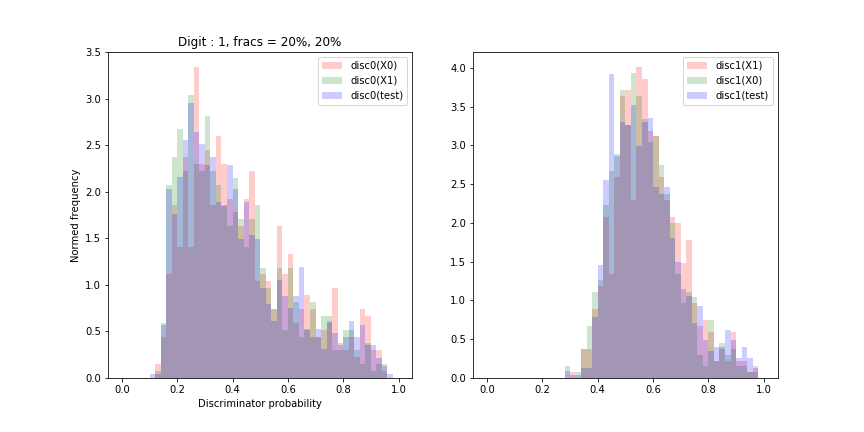}
  \includegraphics[width=0.4\textwidth]{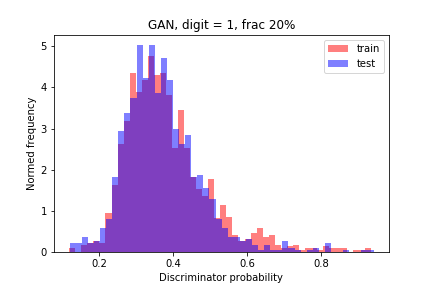}
\end{subfigure}
\caption{\em Same as Fig. \ref{fig:disc_dist_digit0} but for digit 1.
The GAN distributions are very similar, a sign of privacy (see digit 1 in Fig. \ref{fig:wbattacks}).
This distributions for GANs happens only for digit 1.}
\label{fig:disc_dist_digit1}
\end{figure}

Fig. \ref{fig:wbattacks} shows the accuracies of white-box attacks on the discriminators. It shows that privGAN's discriminators are very secure since the attack successes are always close to random guessing while GANs' vulnerability increase as function of training samples.
PrivGAN privacy can also be seen in Fig. \ref{fig:disc_dist_digit0} where the discriminator's confidences on the training set is very similar to the confidence on the holdout set.
For the GAN, on the other hand, we can see a clear distinction between the confidence on the training set and the holdout set.
This is the vulnerability exploited by the white-box attack.
We also see from the GAN confidence distributions, that the confidence decreases on the training set as the training set size decreases.
This is the opposite for the holdout set such that both confidence distributions overlap for small training set size making the synthetic data more private as was shown in Fig. \ref{fig:wbattacks}.

For the bank analogy, this means that an adversary cannot confidently match a bank clients' signatures with its bank by inspecting privGAN components.
This is usually not the case for the GANs.

We show the results of the first four digits to demonstrate the  particularity of digit 1 (the digits not shown have behaviour similar to digits 0, 2 and 3).
Indeed, we note that for digit 1, the GAN discriminator happens to be private too. This can also be seen by the discriminator distributions in Fig. \ref{fig:disc_dist_digit1} where discriminator has a similar response on the training and test samples.
We hypothesise that digits 1s are inherently more private because of their lower diversity, i.e. there are fewer ways to draw 1s than there are for other digits.
As a results,the 1s are more similar and harder to distinguish, hence they are more private.

\section{Conclusion}
We have shown that the utility of a synthetic dataset can be significantly improved for data owners with small datasets when information is shared through the privGAN architecture.
For example, if the data owner has too few data samples, a simple GAN could not even generate recognizable synthetic images while it was always possible with privGAN.
We have tested the most privacy-vulnerable components of the privGAN architecture, the discriminators, and the membership inference attacks are as good as random as opposes to basic GAN synthetic data.
If the data size is large enough, the benefits, besides helping other data owners, are mainly the added privacy to the synthetic data.
Our results are encouraging for enterprises wanting to collaborate by sharing information while minimizing the privacy risks.

We have used a bank example as an application for the privGAN method to preserve the clients privacy while sharing information about their signatures.
In this scenario, a bank (or branch of a bank) could help another bank to generate a synthetic signature dataset to train a better fraudulent signature detection model while preserving their clients' privacy.
We have shown that a bank with too few signatures cannot generated any synthetic signatures on its own, but it becomes possible with the help of another bank through the privGAN architecture.
Depending on the bank needs and privacy policies, it is possible to vary the trade-off between utility and privacy.
Thus, a Pareto front could be explored by modifying certain elements of the privGAN architecture.
For example, the strength of central discriminator \textbf{D\textsubscript{P}} can be modified and will affect both privacy and utility.
It is also possible to change the neural network model to, say, convolutional neural networks which are more adapted to image processing than dense layer networks used here.
This modification might affect the privacy vulnerability as it varies for different types of networks as demonstrated in \cite{chen2019ganleaks}.

The original privGAN paper demonstrated the utility of their methods on more complex datasets such as CIFAR10.
Although it remains to be shown, it is reasonable to assume that our conclusion would also hold with more complex datasets and other data types, within the known limitations of simple GANs.

As for further work, we are exploring the benefits when the data owners have non-i.i.d. subsets.
For example, the signatures could come from clients of different age, gender or any other population parameter that could influence signatures.

\section*{Acknowledgment}
We are grateful to the privGAN authors Sumit Mukherjee, Yixi Xu, Anusua Trivedi and Juan Lavista Ferres for fruitful discussions and granting access to their code.

\bibliographystyle{./bibliography/IEEEtran}
\bibliography{./bibliography/IEEEabrv,./bibliography/IEEEexample}

\end{document}